\documentclass[12pt]{article}
\usepackage{amsmath,amssymb,amsfonts}
\usepackage{graphicx}
\usepackage{xcolor}
\usepackage{hyperref}
\usepackage{caption}
\usepackage{subcaption}
\usepackage{float}
\usepackage{mathrsfs}
\usepackage{cite}
\usepackage{url}
\usepackage{microtype}
\usepackage{booktabs}
\usepackage{authblk}
\usepackage{enumitem}
\usepackage{algorithm}
\usepackage{algorithmic}

\usepackage[margin=1in]{geometry}

\title{Hamiltonian Neural Networks for Robust Out-of-Time Credit Scoring}

\author{Javier Mar\'in}

\date{January 3, 2025}

\begin{document}

\maketitle

\begin{abstract}
This paper presents a novel credit scoring approach using neural networks  to address class imbalance and out-of-time prediction challenges. We develop a specific optimizer and loss function inspired by Hamiltonian mechanics that better captures credit risk dynamics. Testing on the Freddie Mac Single-Family Loan-Level Dataset shows our model achieves superior discriminative power (AUC) in out-of-time scenarios compared to conventional methods. The approach has consistent performance between in-sample and future test sets, maintaining reliability across time periods. This interdisciplinary method spans physical systems theory and financial risk management, offering practical advantages for long-term model stability.

\vspace{0.5em}
\end{abstract}

\section{Introduction}

Credit scoring models have significant challenges with maintaining predicted accuracy over time, primarily because of fluctuating economic conditions and changing data distributions \cite{Baesens2016}. Conventional credit-scoring models have been assessed in static environments, using randomly partitioned datasets for training and validation \cite{Garcia2015}. This approach overlooks the shifting nature of financial data and the possibility of concept drift in credit risk factors \cite{Thomas2002}. The importance of out-of-time (OOT) testing in financial models is critical. Out-of-Time (OOT) testing uses a forward-looking methodology in which adjacent, non-overlapping sample intervals assess model accuracy \cite{Sousa2016}. This approach guarantees that models are evaluated on data representative of future conditions, determinant when measuring a model's prediction accuracy in practical situations \cite{Kim2019}.

OOT hasn't always been effectively used in credit scoring research, mostly due to the lack of long-term datasets \cite{Mushava2024}. The Freddie Mac Single-Family Loan-Level Dataset (SFLLD) fills this void by offering comprehensive historical data, facilitating thorough out-of-time testing \cite{Mushava2024}. Using data from multiple periods for training and testing enables researchers to more accurately evaluate a model's robustness to fluctuations in economic conditions and consumer behaviour \cite{Kvamme2018}. Moreover, credit scoring models must address class imbalance, as instances of default are generally few relative to non-default situations \cite{He2009}. This disparity may result in biased models that underperform on minority classes, requiring particular methodologies for model building and assessment \cite{Lessmann2015}.

Addressing these difficulties requires novel approaches that can effectively represent the temporal dynamics of credit risk while preserving robustness against class imbalance and concept drift. Our proposed Hamiltonian-inspired neural network attempts to address these challenges by using principles from physics to capture the complex, time-dependent nature of credit risk.

\section{Hamiltonian Mechanics in Neural Networks optimization}

Optimization of Neural Networks (NN) is essentially a high-dimensional non-convex optimization problem \cite{Goodfellow2016}. Existing optimization techniques for neural networks, including Stochastic Gradient Descent (SGD) and its derivatives, seek to minimize a loss function through iterative updates of model parameters \cite{Ruder2016}. Yet these techniques frequently face constraints such as slow convergence, trapping in local minima, and difficulties in avoiding saddle points \cite{Dauphin2014}. Recent developments in neural network optimization focused on the geometry of the loss landscape \cite{Li2018} and the dynamics of the optimization process \cite{Yang2019}. The research findings have shown that the optimization trajectory of neural networks can be perceived as a dynamical system, thereby facilitating the application of concepts from physics and dynamical systems theory to neural network optimization \cite{Weinan2017}. A particularly promising approach stems from Hamiltonian mechanics, a reformulation of classical mechanics that offers a solid foundation for the analysis of dynamical systems \cite{Betancourt2017}. The Hamiltonian formalism presents the notion of phase space, a key structure in classical mechanics and current optimization theory \cite{Marsden2013}.

The Hamiltonian formalism provides an effective mathematical framework for developing conservative mechanical system theory, and is a geometric language for multiple fields of physics \cite{Easton1993}. The Hamiltonian formalism introduces phase space, a 2n-dimensional space with $n$ degrees of freedom. Each point in phase space represents an individual state, described by its position and momentum coordinates $(q, p)$. The equations of motion in a Hamiltonian system $H: \mathbb{R}^d \times \mathbb{R}^d \rightarrow \mathbb{R}$ can be defined as
\begin{align}
\frac{dq}{dt} &= \frac{\partial H}{\partial p} \\
\frac{dp}{dt} &= -\frac{\partial H}{\partial q}
\end{align}

In many real-world situations, these equations can't be solved analytically. Symplectic integrators are calculation methods specifically formulated for solving Hamiltonian systems while preserving their fundamental geometric features, particularly the symplectic structure. Symplectic structures are fundamental geometric objects in differential geometry and classical mechanics, and support Hamilton's equations of motion by explaining the connection between position and momentum in physical systems \cite{Marsden2013}. In simple terms, symplectic structures are specific rules that define how things move in physics, similar to an equation for motion. We can define a Hamiltonian in an optimization space system as:
\begin{equation}
H(q, p) = T(p) - V(q)
\end{equation}

Where $q$ represents the current state of the model parameters $(\theta)$, analogous to position in mechanical systems, and $p$ is the difference between consecutive parameter states, $p_t = \theta_{t+1} - \theta_t$, and is analogous to the momentum vector in mechanical systems. In equation (2), $T(p)$ is the 'kinetic energy' term representing the cost of changing the model parameters, and $V(q)$ is the 'potential energy' term representing the loss function of the current model state \cite{Betancourt2017, DeLeon2011, Weinan2017}.

\begin{figure}[h]
    \centering
    \includegraphics[width=0.99\textwidth]{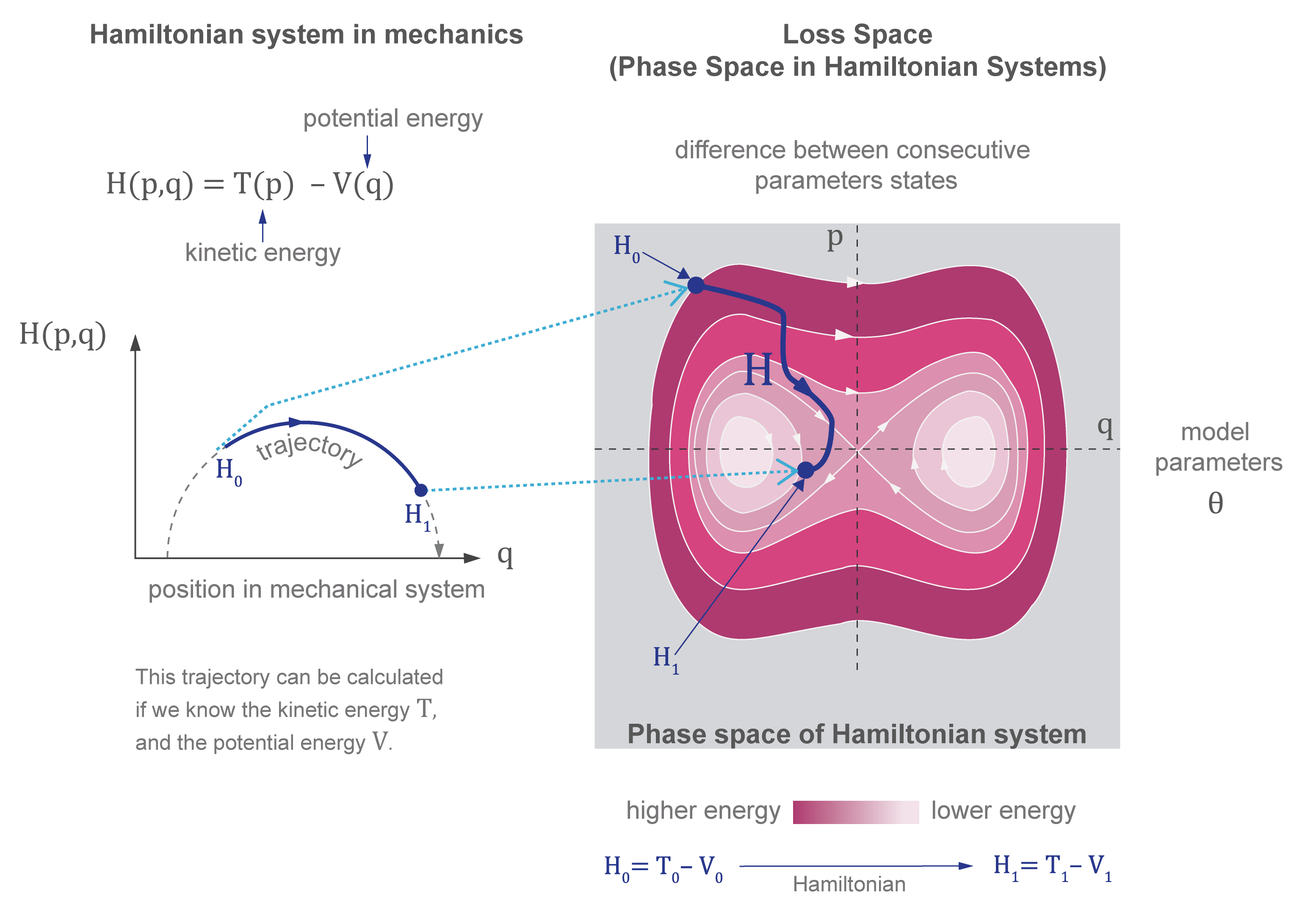}
    \caption{Equivalence between Hamiltonian systems in physical systems, and its application in NN optimization spaces. Position $q$ is equivalent to the NN parameters $\theta$, and momentum vector $p$ is equivalent to the difference between consecutive parameters states.}
    \label{fig:hamiltonian_equivalence}
\end{figure}

Symplectic integrators are numerical methods specifically formulated to solve Hamiltonian systems while preserving their fundamental geometric features, particularly the symplectic structure \cite{Hairer2006}. These integrators are needed for long-term simulations of Hamiltonian systems as they preserve important physical quantities, including energy and angular momentum, with minimal error over long intervals \cite{Marsden2001}. In Hamiltonian systems, a symplectic integrator approximates the flow $\varphi_t$ of Hamilton's system defined by equations (1). Symplectic integrators are characterized by their preservation of the symplectic two-form $\omega = \sum dq_i \wedge dp_i$, which corresponds to the conservation of phase space volume in Hamiltonian systems \cite{Sanz-Serna1992}. This property guarantees the preservation of the qualitative features of the original continuous dynamical system, even with long step sizes \cite{Sanz-Serna1992}. Some examples of integrators are the Störmer-Verlet method, commonly referred to as the 'leapfrog method' in molecular dynamics simulations, is a frequently used symplectic integrator \cite{Verlet1967}. Fourth-order symplectic integrators, like the Forest-Ruth algorithm, offer enhanced long-term stability along with better energy conservation properties \cite{Forest1990}. One of the simplest forms of symplectic integration is the first-order symplectic integrator called the Symplectic Euler method \cite{Leimkuhler2004}. For a Hamiltonian system with position $q$ and momentum $p$, the Symplectic Euler method can be represented as:
\begin{align}
p_{t+1} &= p_t - \Delta t \times \frac{\partial H}{\partial q}(q_t, p_{t+1}) \\
q_{t+1} &= q_t - \Delta t \times \frac{\partial H}{\partial p}(q_t, p_{t+1})
\end{align}

where $H$ is the Hamiltonian and $\Delta t$ is the time step, analogous to the learning rate in a non-convex optimization problem. In this paper we proposed a hybrid approach that combines the basic structure of Symplectic Euler method, momentum-based updates \cite{Sutskever2013}, and a Hamiltonian-inspired normalization factor. This combination aims to use the advantages of each component: the preservation of symplectic structure inherent in the Euler method, the better convergence features of momentum-based methods, and the energy conservation principles derived from Hamiltonian mechanics \cite{Ma2015}. Our hybrid approach use a momentum update similar to Adam or RMSprop optimizers. We can represent it as:
\begin{equation}
m_t = \beta \times m_{(t-1)} + (1 - \beta) \times g_t
\end{equation}

Where $m$ is the momentum, $\beta$ is a decay factor, and $g_t$ is the gradient. Our parameter update includes a Hamiltonian normalization:
\begin{equation}
\theta_t = \theta_{t-1} - \frac{\eta \times m_t}{\sqrt{H_t + \epsilon}}
\end{equation}

Then we compute the Hamiltonian as the sum of kinetic and potential energy $H(q, p) = T(p) - V(q)$. Our approach preserves the idea of symplectic integration by separating momentum (7), and position (8) updates. These approaches have been shown to accelerate convergence and help overcome local minima in neural network training \cite{Sutskever2013}. The normalization by $\sqrt{H_t + \epsilon}$ in (8) is special, and not part of standard symplectic integrators ($\epsilon$ is a small constant for stability). This is derived from the energy conservation principles in Hamiltonian mechanics, and aids in adaptively adjusting the step size based on the "energy" of the system, potentially leading to more stable and efficient optimization \cite{Ma2015}. The momentum decay factor $\beta$, is borrowed from modern optimization techniques.

\begin{algorithm}[h]
\caption{Symplectic Optimizer}
\begin{minipage}{\textwidth}
\begin{algorithmic}
\REQUIRE Learning rate $\eta$, momentum coefficient $\beta$, epsilon $\epsilon$
\FOR{each iteration}
    \FOR{each parameter $\theta$ in model parameters}
        \STATE $g \leftarrow \nabla L(\theta)$ \hfill // Compute gradient
        \STATE $v \leftarrow \text{state}[\theta]['momentum']$ \hfill // momentum
        \STATE // Update momentum
        \STATE $v \leftarrow \beta \times v + (1 - \beta) \times g$
        \STATE // Compute Hamiltonian
        \STATE $K \leftarrow 0.5 * ||v||^2$ \hfill // Kinetic energy
        \STATE $V \leftarrow 0.5 * ||\theta||^2$ \hfill // Potential energy
        \STATE $H \leftarrow K + V$ \hfill // Hamiltonian
        \STATE // Update parameter
        \STATE $\theta \leftarrow \theta - \eta * v / (\sqrt{H + \epsilon})$
        \STATE // Store updated momentum
        \STATE $\text{state}[\theta]['momentum'] \leftarrow v$
    \ENDFOR
\ENDFOR
\end{algorithmic}
\end{minipage}
\end{algorithm}

The advantages of our proposal are the following: better energy conservation, improved long-term stability, and potential for larger learning rates. Conventional loss functions in neural networks, such as cross-entropy, primarily aim to minimize prediction errors \cite{Goodfellow2016}. In the complex optimization landscapes characteristic of deep learning, these functions may fail to accurately represent the geometrical properties of the parameter space or enable stable learning dynamics \cite{Li2018}. Our Hamiltonian-inspired loss function covers these limitations by integrating principles from classical mechanics.

We define the loss function as:
\begin{equation}
H_{loss}(\theta) = L_{base}(\theta) + \lambda \times R(\theta)
\end{equation}

where $L_{base}$ represents the standard loss (e.g., cross-entropy), $R(\theta)$ denotes a regularization term analogous to potential energy in Hamiltonian systems, and $\lambda$ is a hyperparameter that controls the strength of regularization \cite{Betancourt2017}. The regularization term $R(\theta) = (1/2)||\theta||^2$ penalizes large parameter values, leading to more accurate solutions. This formulation is derived from the potential energy component in Hamiltonian mechanics, $V(q) = (1/2)||p||^2$, where $p$ denotes momentum \cite{Marsden2013}.

Through the incorporation of this physics-inspired regularization, our loss function seeks to enhance generalization by promoting more uniform parameter distributions \cite{Neyshabur2017}. At the same time, we can improve optimization stability by applying the energy conservation principles inherent in Hamiltonian systems \cite{Weinan2017}. With this approach, we provide a natural framework for balancing model complexity and data fit, analogous to the connection between kinetic and potential energy in physical systems.

\section{Data and Methodology}

We use the FMData dataset introduced by Mushava \& Murray \cite{Mushava2024}. This dataset originates from the Freddie Mac Single-Family Loan-Level Dataset (SFLLD) and includes mortgage loans granted in the USA, along with their subsequent repayment patterns. It contains a list of fully amortized fixed-rate mortgage loans. The data covers the period from 1999 to a more recent date, based on the access date, and includes both loan origination data and monthly performance metrics. A unique loan identifier is used to gather loan origination and repayment information. Default is represented by the failure to make three or more payments within defined performance intervals of 12, 24, 36, 48, and 60 months.

In our experiment, we use the out-of-time (OOT) dataset from FMData, intended to evaluate model performance on future data through non-overlapping temporal intervals for training and testing. The OOT test set originates from a period following to training set data. It attempts to recreate real-world scenarios in which models forecast future, unobserved data. This dataset includes performance windows of 12, 36, and 60 months. This data offers a practical credit scoring scenario, enhancing the relevance of our research to real financial risk analysis. The OOT dataset enables the assessment of our model's performance over time, which is essential in credit scoring due to fluctuating economic conditions. Using out-of-time data evaluates our model's capacity to generalize to future, unobserved data – an essential criterion in practical applications. The data from Freddie Mac is highly valued in the financial sector, increasing the credibility of our research.

Our experimental setup is designed to carefully evaluate the effectiveness of our Hamiltonian-inspired neural network methodology in credit scoring, with particular emphasis on how well it holds up in out-of-time (OOT) contexts. The configuration includes data partitioning, management of class imbalance, hyperparameter tuning, and a cross-validation approach. We split the FM Data into three sets: a training set for model training that includes loans from the earliest time frame, a validation set for hyperparameter optimization and early termination of training, and a test set with loans from a subsequent time frame to ensure no temporal overlap. The OOT test set is essential for evaluating the model's capacity to generalize to future, unseen data, simulating real-world credit scoring situations \cite{Kvamme2018}.

Credit default datasets often exhibit considerable class imbalance \cite{Lessmann2015}. We implement the Synthetic Minority Over-sampling Technique (SMOTE) \cite{Chawla2002} on the training set to resolve this limitation. SMOTE generates synthetic instances of the minority class (defaults) within the feature space. It's very important that we only implement SMOTE on the training data, thereby ensuring that the validation and out-of-time test sets retain their original distributions, thus preserving the integrity of our evaluation \cite{DalPozzolo2015}.

We use a grid search methodology for hyperparameter optimization \cite{Bergstra2012}. The principal hyperparameters adjusted include: Learning rate ($\eta$), Momentum coefficient ($\beta$), Regularization intensity ($\lambda$), and network architecture (quantity and dimensions of hidden layers). Optimization involves using the validation set to prevent overfitting to the training data. We use a time-based cross-validation strategy to guarantee accurate performance estimation \cite{Arlot2010}. This methodology identifies the temporal patterns of the data. The training data is partitioned into $K$ folds while preserving temporal sequence. For each fold $k$, we train on folds 1 through $k$ and validate on fold $k+1$. The procedure is executed $K-1$ times, with the final fold designated for conclusive validation. This strategy offers a more accurate evaluation of the model's performance over time, essential for credit scoring applications.

We perform a comparative analysis to benchmark the performance of our Hamiltonian-inspired neural network against XGBoost, an effective gradient boosting algorithm commonly used in credit scoring applications \cite{Chen2016}. XGBoost has shown remarkable efficiency in many machine learning applications, especially in structured data problems such as credit risk evaluation \cite{Brown2012}. We use XGBoost applying the same data partitioning, class imbalance management, and cross-validation methodology as our Hamiltonian-inspired model. This guarantees a fair comparison between the two approaches \cite{Xia2017}. The XGBoost model is optimized using its specific hyperparameters, which include: number of estimators, maximum tree depth, learning rate, subsample ratio, and column sample ratio. Hyperparameter optimization for XGBoost is achieved through randomized search with cross-validation, a technique proven to be effective for complex models that have many hyperparameters \cite{Bergstra2013}.

We propose to evaluate the performance of our Hamiltonian-inspired method in comparison to XGBoost, an established benchmark in credit scoring \cite{Lessmann2015}. Evaluating the robustness of both models in out-of-time scenarios is essential for effective credit risk assessment, as well as analyzing the trade-offs among model complexity, interpretability, and predictive accuracy in the field of credit scoring \cite{Fitzpatrick2016}.

\section{Results and discussion}

The first part of the experiment has been to find the best hyperparameters for both algorithms. In table \ref{tab:hyperparameters} we can see the best hyperparameters found for both algorithms.

\begin{table}[h]
\centering
\caption{Best hyperparameters found for experiments}
\label{tab:hyperparameters}
\begin{tabular}{lc}
\toprule
\multicolumn{2}{c}{Neural Network with Hamiltonian approach} \\
\midrule
Regulation coefficient, $\lambda$ & 0.01 \\
Learning rate & 0.01 \\
Hidden dimensions & [128, 64] \\
Dropout rate & 0.2 \\
Activation function & Leaky ReLu \\
\midrule
\multicolumn{2}{c}{XGBoost algorithm} \\
\midrule
Learning rate & 0.3 \\
Max depth & 7 \\
Min child weight & 1 \\
Number of estimators & 300 \\
\bottomrule
\end{tabular}
\end{table}

Table \ref{tab:comparative-results} presents the comparative results of our Hamiltonian-inspired methodology and the XGBoost algorithm across three different Out-of-Time (OOT) datasets (FM12, FM36, and FM60). XGBoost consistently displays higher accuracy, precision, recall, and F1 scores across all temporal horizons. In the FM12 dataset, XGBoost achieves an accuracy of 0.9871, whereas the Hamiltonian approach scores 0.8047. This trend persists in FM36 and FM60, suggesting XGBoost's improved results in more common binary classification metrics.

\begin{table}[h]
\centering
\caption{Comparative results}
\label{tab:comparative-results}
\begin{tabular}{lccc}
\toprule
OOT dataset & FM12 (12 months) & FM36 (36 months) & FM60 (60 months) \\
\midrule
\multicolumn{4}{c}{Hamiltonian inspired approach} \\
\midrule
Mean accuracy & 0.8047 $\mp$0.0021 & 0.7638 $\mp$0.0010 & 0.6975 $\mp$0.0003 \\
Mean F1 Score & 0.8012 $\mp$0.0000 & 0.7608 $\mp$0.0008 & 0.6968 $\mp$0.0003 \\
Mean AUC score & 0.8027 $\mp$0.0000 & 0.764 $\mp$0.0000 & 0.6974 $\mp$0.0000 \\
Mean Precision & 0.8201 $\mp$0.0000 & 0.777 $\mp$0.0000 & 0.6994 $\mp$0.0000 \\
Mean Recall & 0.8381 $\mp$0.0001 & 0.7638 $\mp$0.0000 & 0.6973 $\mp$0.0000 \\
\midrule
\multicolumn{4}{c}{XGBoost algorithm} \\
\midrule
Accuracy & 0.9871 & 0.9663 & 0.9319 \\
Precision & 0.9948 & 0.9831 & 0.9796 \\
Recall & 0.9827 & 0.9663 & 0.9319 \\
F1 Score & 0.9882 & 0.9745 & 0.9542 \\
AUC & 0.6072 & 0.6221 & 0.6665 \\
\bottomrule
\end{tabular}
\end{table}

The Hamiltonian-inspired method offers markedly superior Area Under the Curve (AUC) scores across all datasets. FM12 reaches an AUC of 0.8027, whereas XGBoost achieves 0.6072. This pattern is uniform in FM36 (0.764 compared to 0.6221) and FM60 (0.6974 compared to 0.6665). Higher AUC scores indicate that our approach has better discriminative capability and ranking competence, which is critical in credit scoring applications \cite{Hand2013}. The AUC values calculated by the XGBoost algorithm are close to 0.5, suggesting that, despite its accuracy, its capacity for discrimination in unseen data is very close to randomness. Our approach implies more consistency in performance across different time horizons. The fluctuation in its metrics from FM12 to FM60 is minimal, with AUC scores varying from 0.8027 to 0.6974. XGBoost demonstrates superior initial performance; however, it indicates a significant decline in accuracy and recall over extended time periods. XGBoost has consistent high precision across all datasets (ranging from 0.9948 to 0.9796), yet demonstrates a gradual decline in recall over time (from 0.9827 to 0.9319). The Hamiltonian approach, despite exhibiting lower absolute values, reveals a more balanced precision-recall trade-off, which is important in the context of imbalanced datasets frequently found in credit scoring \cite{Sousa2016}. The consistent stability of the metrics in the Hamiltonian approach across different time windows suggests better generalization to future, unseen data. This part is particularly important for credit risk evaluation, as model performance typically declines over time due to concept drift \cite{Sousa2016}.

Although XGBoost excels in standard metrics, the higher AUC scores of the Hamiltonian method suggest its potential advantage in ranking borrowers by risk. This is critical in practical credit scoring problems, where the rank-ordering of risk frequently holds greater significance than binary classification \cite{Verbraken2014}.

In summary, although XGBoost demonstrates exceptional performance in conventional classification metrics, our Hamiltonian-inspired methodology reveals promising results for discriminative power (AUC) and temporal stability. The findings indicate that our approach may provide benefits in situations where accurate risk ranking and persistent model stability have an edge over basic classification accuracy. Further investigation of the model's performance across different data subsets and its interpretability could provide further insights into its practical value in credit risk rating.

\section{Conclusions}

Our research introduces a Hamiltonian-inspired neural network for credit scoring that shows promising results in out-of-time prediction scenarios. Although our approach is not superior to XGBoost in conventional accuracy metrics, it shows a clear potential in critical aspects of real credit risk assessment, specifically in discriminative power and temporal stability.

The higher AUC scores reached by our approach across different time periods suggest an improved capacity to rank credit risks effectively. This capability is particularly important in the finance industry, where accurate risk stratification frequently outweighs binary classification. Furthermore, the consistent performance of our model across different time frames demonstrates its resilience to temporal shifts, a significant quality in the dynamic field of credit risk. Nonetheless, we should look at these outcomes with moderate optimism. The consistently superior accuracy, precision, and recall metrics of XGBoost highlight the enduring efficiency of traditional machine learning approaches in credit scoring. Our Hamiltonian-inspired approach, although promising, requires additional refinement and validation prior to its consideration for practical implementation. Exploring the integration of our Hamiltonian-inspired methodology with conventional techniques like XGBoost could produce a model that leverages the strengths of both approaches. Improving the interpretability of our model is essential for its acceptance in the strongly regulated financial sector. An in-depth examination of our model's robustness against concept drift may yield significant insights for persistent model stability. Analyzing how our methodology identifies and uses feature importance may provide novel insights into credit risk drivers.

The financial industry stands at a critical juncture, facing increasingly complex risk landscapes and regulatory pressures. The pursuit of novel, more robust approaches to credit scoring is not merely an academic exercise but a necessity for maintaining financial stability and fostering economic growth. Although our findings are promising, they underscore the importance of ongoing innovation and thorough assessment in this area. A collaboration between academia and industry will be key to the development and validation of robust methodologies that can adapt to market dynamics.

In conclusion, our research enriches the discussion on improving credit scoring methodologies. It highlights the value of interdisciplinary methods in addressing complex financial problems. We aim to improve and expand this work to improve credit risk assessment, thereby seeking a more stable and efficient financial ecosystem.

\bibliographystyle{plainnat}

\end{document}